\documentclass{isprs} 
\usepackage{amsmath,amssymb,amsfonts}
\usepackage{calrsfs}
\DeclareMathAlphabet{\pazocal}{OMS}{zplm}{m}{n}
\usepackage{cite}
\usepackage{graphicx}
\usepackage{mathrsfs}
\usepackage{xcolor}
\usepackage{balance}
\usepackage{hyperref}
\usepackage{algorithm}
\usepackage[noend]{algpseudocode}

\usepackage{subfigure}
\usepackage{setspace}
\usepackage{geometry} 
\usepackage{epstopdf}
\usepackage[labelsep=period]{caption}  
\usepackage[british]{babel} 
\usepackage[hang]{footmisc}

\begin{document}

\title{UAS Visual Navigation in Large and Unseen Environments via a Meta Agent}
\date{}


\author{
    Yuci Han\textsuperscript{a}, Charles Toth\textsuperscript{b},
    Alper Yilmaz\textsuperscript{a}
    }

\address
{
	\textsuperscript{a } Photogrammetric Computer Vision Laboratory,\\
\textsuperscript{b} Satellite Positioning and Inertial Navigation Laboratory\\
	The Ohio State University, Columbus, OH 43210, USA \\ 
	\{han.1489, toth.2, yilmaz.15\}@osu.edu
}
\abstract{

The aim of this work is to develop an approach that enables Unmanned Aerial System (UAS) to efficiently learn to navigate in large-scale urban environments and transfer their acquired expertise to novel environments. To achieve this, we propose a meta-curriculum training scheme. First, meta-training allows the agent to learn a master policy to generalize across tasks. The resulting model is then fine-tuned on the downstream tasks. We organize the training curriculum in a hierarchical manner such that the agent is guided from coarse to fine towards the target task. In addition, we introduce Incremental Self-Adaptive Reinforcement learning (ISAR), an algorithm that combines the ideas of incremental learning and meta-reinforcement learning (MRL). In contrast to traditional reinforcement learning (RL), which focuses on acquiring a policy for a specific task, MRL aims to learn a policy with fast transfer ability to novel tasks. However, the MRL training process is time consuming, whereas our proposed ISAR algorithm achieves faster convergence than the conventional MRL algorithm. We evaluate the proposed methodologies in simulated environments and demonstrate that using this training philosophy in conjunction with the ISAR algorithm significantly improves the convergence speed for navigation in large-scale cities and the adaptation proficiency in novel environments. The project page is publicly available at \url{https://superhan2611.github.io/isar_nav/}.

}

\keywords{Visual Navigation, Deep Reinforcement Learning, Unmanned Aerial System.}

\maketitle

\section{Introduction}

Long-range navigation is a complex cognitive task that relies on robust localization techniques using GPS,  as well as a variety of subtasks, such as detection and mapping. This work investigates autonomous navigation using only visual observations from a single monocular camera, mimicking the way that humans would navigate in an unfamiliar environment without GPS \cite{lee2015neural} \cite{Wang2018PrefrontalCA}. Visual navigation has potential for practical applications in scenarios where other sensors are inaccessible.

With the development of deep reinforcement learning (DRL), various end-to-end DRL approaches have been developed for visual autonomous navigation tasks in small-scale indoor environments \cite{zhu2017icra} \cite{8793493} \cite{Wang} \cite{8953608} \cite{Han} \cite{9967103} \cite{Hong2020VLNBERTAR}. However, long-range visual navigation in urban environments is significantly more challenging than indoor environments and has not yet been fully explored. 
In particular, urban environments are characterized by their complexity, diverse scenes, and numerous obstacles. It is difficult to apply traditional DRL training techniques designed for indoor-scale tasks to urban navigation due to data-inefficient and high training costs. In addition, low-altitude UAS navigation allows free flight in open spaces, introducing higher complexity and flexibility that further complicates the challenge. 

\begin{figure}[t!]
\centering
\includegraphics[width=0.9\linewidth]{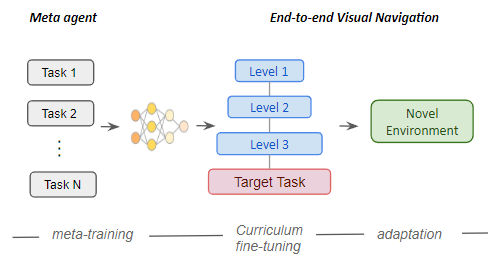}
\caption{\textbf{Overview of learning framework.} The navigation task consists of two phases: meta-training and curriculum fine-tuning. Meta-training allows the agent to learn a master navigation policy. The hierarchical-structured curriculum adapts the meta-policy to the target task. This meta-policy can further be transferred  to novel environments.}
\label{fig1}
\end{figure}

In this paper, our aim is to develop an approach for learning to navigate on a large scale in less time and to adapt quickly to new environments. To this end, we introduce a meta-curriculum training scheme. Inspired by meta-learning theory \cite{Finn2017ModelAgnosticMF} \cite{Nichol2018ReptileAS} \cite{Wortsman2018LearningTL}, the proposed training strategy engages in `learn how to learn' and `learn to navigate'. More concretely, we decompose the training into two phases: meta-training and curriculum fine-tuning (see Fig. \ref{fig1}). Meta-training allows the agent to develop essential navigation skills by exploring a variety of small tasks. This prior knowledge can then be transferred to new tasks, enabling fast adaptation. In the next phase, we fine-tune the meta-agent on the target task using a curriculum training scheme. This strategy increases the complexity of hierarchical training tasks by providing the agent with progressively more difficult environments that help the agent learn to reach more distant destinations. In addition, we propose an incremental self-adaptive reinforcement learning algorithm (ISAR) to accelerate the learning speed of the traditional MRL approach for long-episode tasks with sparse rewards. ISAR learns an interaction loss and an adaptation loss incrementally within an episode, allowing the agent to optimize its policy through self-adaptive exploration, leading to improved learning and exploration efficiency.  

The contributions of this paper are summarized as follows: \textbf{(a)} We present a two-stage framework for training an end-to-end low-altitude UAS visual navigation system in urban environments. Curriculum training improves the efficiency of learning the high-complexity long-range navigation task, and meta-training enables the generalization ability to new environments. \textbf{(b)} We introduce ISAR, an improved MRL algorithm for the long-episode reinforcement learning task. The key components of ISAR are the inner-episode adaptation loss design and the incremental update procedure, which allow the agent to learn to adapt from their own experiences.  \textbf{(c)} We present extensive experimental validation in the AirSim simulation environment \cite{10.1007/978-3-319-67361-5_40}. The results verify the superiority of our training strategy over other traditional reinforcement learning approaches. Additionally, we demonstrate that the ISAR algorithm outperforms the MRL baseline in terms of convergence speed. Furthermore, the meta-training policy shows a notable transfer ability to novel environments.

\section{Related Work}

\noindent \textbf{Deep Reinforcement Learning Navigation.} Reinforcement learning (RL) algorithms have been applied to a wide range of domains, including autonomous driving \cite{9956121}, robotics \cite{Luo2020AFS} \cite{Kahn2017SelfSupervisedDR}, and video games \cite{DBLP}. Recently, RL has increased in popularity for navigation tasks due to its flexibility in learning complex decision-making tasks.  However, most of the published literature is focused on indoor simulated environments \cite{Wu2020TowardsTV} \cite{Shen2019SituationalFO}. Li et al. \cite{Li2019GraphAM} used graph attention memory (GAM) to address the long-term memory limitation of reactive deep reinforcement learning. Kwon et al. \cite{Obin} investigated a similar problem and introduced visual graph memory (VGM) that can learn a goal-oriented policy directly from the graph representation. Wu et al. \cite{Wu2019BayesianRM} improved the generalizability of semantic visual navigation agents by using Bayesian Relational Memory (BRM). However, these models are fixed during the transfer process and do not generalize to new environments. On the other hand, Wortsman et al. \cite{Wortsman2018LearningTL} proposed a self-adaptive visual navigation model that can learn during the training and inference stages. Like this work, we develop a transferable model for the urban navigation task. Unlike this work, we propose a more effective training framework and an ISAR algorithm that achieve faster convergence in large-scale urban navigation and adapt quickly to novel environments.

\noindent \textbf{Meta-Learning.} Meta-learning, also known as `learning to learn,' is a rapidly growing field in machine learning and artificial intelligence \cite{Alex} \cite{rusu2018metalearning} \cite{Mingzhang}. The core idea of meta-learning is to learn algorithms or models that can quickly adapt to new tasks or environments with a small amount of data. Gradient-based meta-learning \cite{Hochreiter2001LearningTL} \cite{frans2018meta} \cite{Ravi2016OptimizationAA} \cite{Alex} has shown promising results in various domains, especially for the few-shot learning problem \cite{DBLP:journals/corr/LiZCL17} \cite{8954011}. The basic idea behind gradient-based meta-learning is to optimize the model's parameters with the gradients of the loss function over a set of tasks so that it can quickly adapt to new tasks. Andrychowicz et al. \cite{Andrychowicz2016LearningTL} proposed a meta-learning approach that learns to learn by optimizing the learning algorithm itself and achieves state-of-the-art results on several meta-learning benchmarks. Lee et al. \cite{8954109} proposed MetaOptNet, which learns feature embedding that generalizes well under a linear classification rule for novel categories. Finn et al. \cite{Finn2017ModelAgnosticMF} proposed a meta-learning approach called Model-Agnostic Meta-Learning (MAML) that can be applied to any differentiable model or network architecture. Li et al. \cite{Li2017LearningTG} and Yu et al. \cite{DBLP:conf/rss/YuFDXZAL18} augmented the MAML algorithm by synthesizing virtual testing domains and simulating domain shift during training so that it transfers supervision in one domain to another. Our proposed training scheme relies on the MAML model during the meta-training stage. Furthermore, our ISAR algorithm integrates meta-learning and incremental learning into a per-episode exploration process of the conventional MRL algorithm, which improves the learning and exploration efficiency.

\section{Methodology}

\subsection{Urban Visual Navigation Task}

\begin{figure}[t!]
\centering
\includegraphics[width=\linewidth]{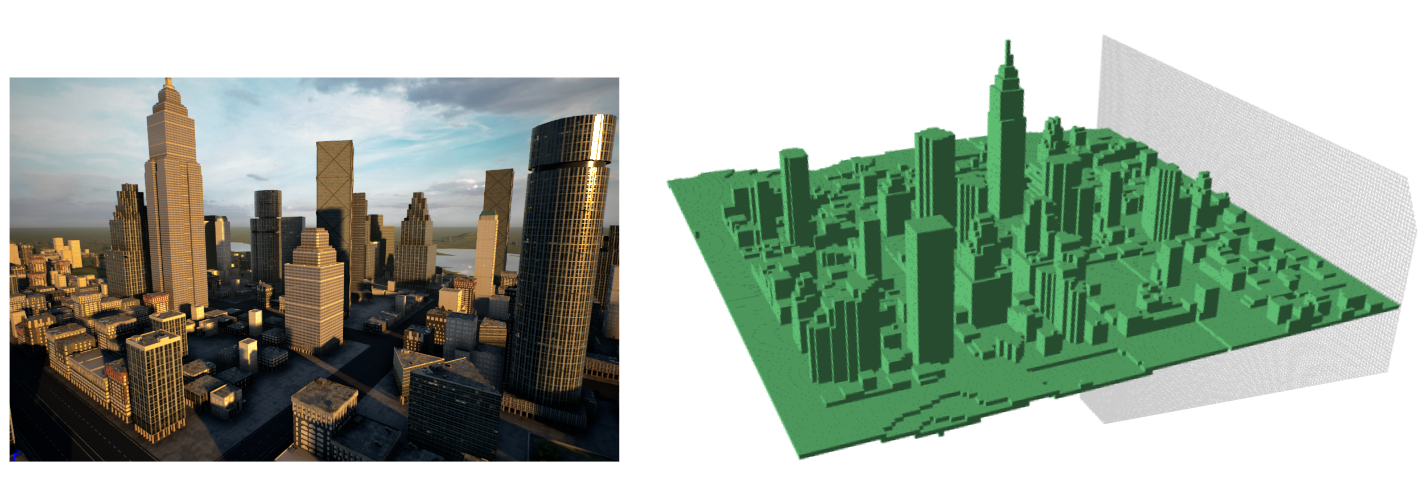}
\caption{The illustration of AirSim simulated urban environment.}
\label{fig2}
\end{figure}

The objective of the visual navigation task of the UAS is to train an autonomous agent to navigate from an arbitrary starting point to the designated destination. For this project, we use AirSim, a simulated environment that consists of diverse realistic scenes and numerous objects, as the deployment platform \cite{airsim} (see Fig. \ref{fig2}). 

The objectives of this work are to: (1) train the UAS agent to perform low-altitude ($15m$) urban navigation; and (2) assess the transfer performance of the meta-policy in a novel environment. We formalize the navigation task as a Markov Decision Process $(\mathcal S, \mathcal A, p, r)$ where $\mathcal S$ is the state space, which consists of only image inputs captured by a downward looking camera; $\mathcal A$ is the action space, which includes \textit{moving forward}, \textit{moving right}, \textit{moving left}, \textit{moving backward}; $p$ is the transition function between states and $r$ is the reward. In the navigation task, the agent starts the episode from a randomly selected initial position and explores the environment to reach the destination in as few steps as possible. During exploration, the agent selects a series of actions from $\mathcal A$, and receives corresponding rewards from $r$ until the episode ends. We assign a reward of $5$ points for successful reaching the destination, otherwise, $-0.01$ points is applied as time penalty.

\subsection{Meta-Curriculum Training Framework}

As illustrated in Fig. \ref{fig1}, the learning process comprises the meta-training stage and the curriculum fine-tuning stage. The meta-training prepares a meta-agent for downstream tasks, and the curriculum fine-tuning adapts the meta-policy to the target task. 

\begin{figure}[t!]
\centering
\includegraphics[width=3.4in]{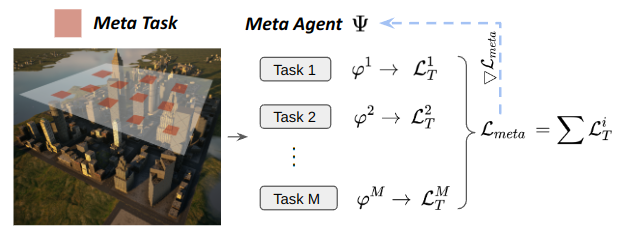}
\caption{The illustration of meta-training process. $\mathcal L_{T}^i$ is the loss for the $i^{th}$ meta-task. The meta-loss $\mathcal L_{meta}$ is the sum of $\mathcal L_{T}^i$, which is used to update the meta-agent $\psi$.}
\label{fig3}
\end{figure}

\noindent \textbf{Meta-Training.}  \space The key idea of meta-training is to enable the agent to build up its understanding of the navigation rules by practicing on smaller tasks that can be mastered quickly and easily. More precisely, we use $g_\psi$ to represent the meta-policy, where $\psi$ is the weight of a deep learning model. We generate $M$ meta-tasks as the training task set $\mathcal T_{meta}$. Each task is defined as $\mathcal T^i=\{\mathcal L_{T}^i(\mathcal S, \mathcal A, \mathcal R), \mathcal E_{T}^i\}$ that consists of task loss $\mathcal L_{T}^i$ and task environment $\mathcal E_{T}^i$. In the context of urban navigation task, we train the meta-agent at an altitude of $300m$ (see Fig. \ref{fig3}). This is because the drone can capture a wider and more comprehensive view of the environment at this level, which benefits its adaptation to lower-altitude navigation. Consequently, as illustrated in Figure \ref{fig3}, $\mathcal E_{T}^i$ is a subregion generated at the altitude of $300m$. Each $\mathcal T^i$ is an indivisual navigation task with a designated target. We draw meta-tasks in batches from $\mathcal T_{meta}$ in sequential order for training. To begin with, we employ the RL algorithm to learn the navigation policy $\pi(\varphi^i)$ with parameters $\varphi^i$ for the meta-task $\mathcal T^i$ and then calculate $\mathcal L_{T}^i$. The meta-policy $\psi$ is trained by optimizing performance across all meta-tasks. Therefore, the meta-loss $\mathcal L_{meta}$ is defined as the sum of all losses from meta-tasks: $\sum \mathcal L_{T}^i$. We utilize stochastic gradient descent (SGD) with a learning rate of $\eta$ to update the meta-agent (see Fig. \ref{fig3}): 
\begin{equation}
\psi' = \psi - \eta \nabla_{\psi} \mathop \sum \limits_{\mathcal T_i \sim \mathcal T_{meta} } \mathcal L_{T}^i
\end{equation}
Following this step, we update the parameters of $\pi(\varphi^i)$ with $\psi'$ and repeat this process until convergence. According to the meta-learning theory \cite{Finn2017ModelAgnosticMF}, this approach encourages the discovery of parameters $\psi$ sensitive to changes such that slight modifications to the parameters lead to significant improvements in new tasks. Therefore, the meta-agent can be fine-tuned through a gradient-based approach, enabling it to make rapid adaptation on novel tasks.

\begin{figure}
\centering
\includegraphics[width=3in]{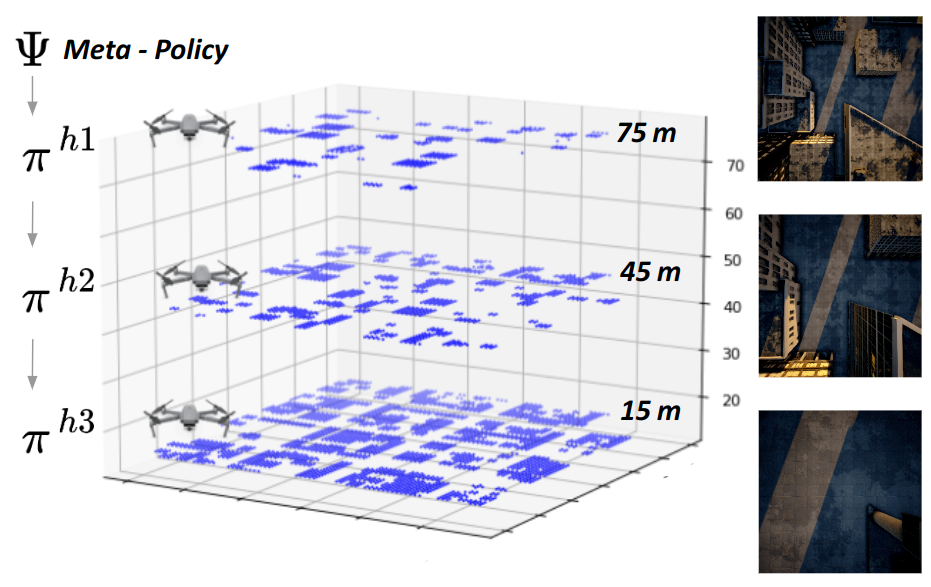}
   \caption{The illustration of the obstacles distribution and the observation of the same location at altitudes of $75m$, $45m$ and $15m$. The agent starts with the meta-policy and learns hierarchical policies from coarse-to-fine.}
   \label{fig4}
\end{figure}

\noindent \textbf{Hierarchical Curriculum Training.} \space In this phase, we fine-tune the meta-policy to the target task. Compared to indoor navigation, urban navigation is significantly more challenging due to its larger scale of environment and more complex scenarios. Additionally, navigating at lower altitudes is even harder than at higher altitudes, as there are numerous obstacles and narrower perspectives (with a downward camera) (see Fig. \ref{fig4}).

Our proposed curriculum training scheme aims to accelerate the learning speed by guiding the agent to learn the policy from `coarse to fine.' As illustrated in Fig. \ref{fig4}, the number of obstacles decreases with higher altitudes. Consequently, our training curriculum is structured hierarchically starting from higher altitudes and gradually moving towards lower altitudes as task difficulty increases. In particular, we start by initializing the agent with a meta-policy and fine-tune it at an altitude of $75$ meters with sparse obstacles. As the agent gradually refines its policy, we subsequently reduce the flight height to $45$ meters. Ultimately, we set the drone's height at $15$ meters to complete the target task.

\subsection{Incremental Self-Adaptive Reinforcement Learning}

 Previous works \cite{Wortsman2018LearningTL} \cite{Luo2020AFS}\ \cite{Li2019UnsupervisedRL} \cite{Mayo2021VisualNW} use meta reinforcement learning for transfer navigation in unseen environments. Generally, MRL can learn a policy with rapid adaptation ability by introducing a meta-loss at the end of the episodes or tasks. Our proposed ISAR algorithm is essentially an optimization of conventional MRL for long-episode task such as large-scale navigation. Like MRL, ISAR can leverage any existing RL algorithms, such as actor-critic (A3C) \cite{Volodymyr}, double Q-learning \cite{Hasselt2010DoubleQ}, or PPO \cite{John}, as the base model. Before formally presenting the ISAR algorithm, we begin with an overview of the traditional reinforcement learning process applied to the navigation task.

We use $s_t$ to denote the agent's state at time $t$, which is an RGB image observed by the drone. The navigation policy $\pi(a_t|s_t, s_{target})$ is a distribution over actions given $s_t$ and the target observation $s_{target}$. In the DRL model, this policy is represented by a network with parameters $\theta$, denoted as $\pi_\theta(a_t|s_t, s_{target})$. We use ResNet18 to extract the features of $s_t$ and $s_{target}$. The combined feature is then fed into the policy network to generate the navigation policy and value. During an exploration episode, the agent starts from $s_0$, takes a sequence of actions according to the policy $\pi_\theta(a_t|s_t, s_{target})$ and receives feedback in the form of rewards $r$ which guides the agent towards the target. Our goal is to maximize the sum of discounted rewards by minimizing the loss $\mathcal{L}_{episode}$ throughout the episode.

\begin{figure*}
\begin{center}
\includegraphics[width=5.7in]{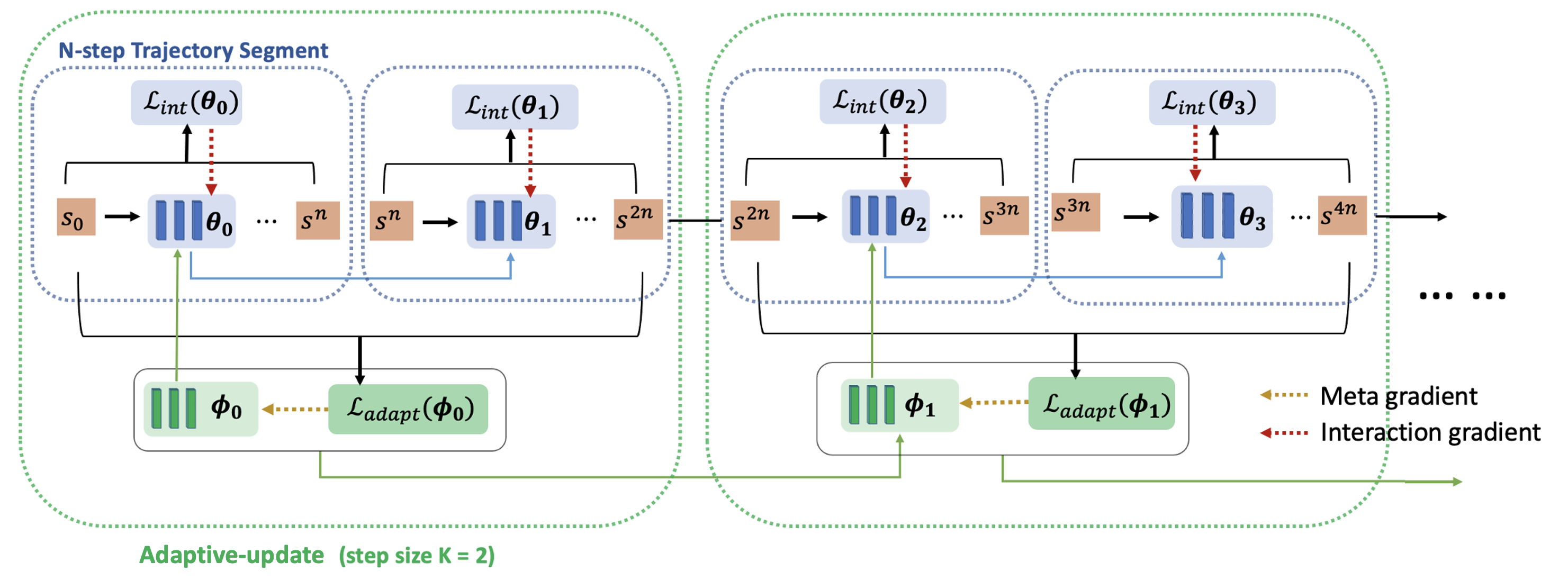}
\end{center}
   \caption{The illustration of the ISAR framework with adaptive-update step size = $2$. During exploration, we learn two types of losses: the interaction loss $\mathcal{L}_{int}$ to update the interaction policy $f_\theta$, and the adaptive loss $\mathcal{L}_{adapt}$ based on trajectory segment with length $N$ to update the adaptation policy $f_\phi$.}
   \label{fig6}
\end{figure*}

The proposed ISAR algorithm accelerates this learning process by incremental self-adaptive exploration (Figure \ref{fig6}). The policy update in traditional MRL is typically performed on episode-basis or task-basis using episodes-loss or tasks-loss as the meta-loss $\mathcal{L}_{meta}$. In our model, we enable the agent to learn the navigation policy incrementally by allowing it to update based on a small number of interactions at a time within the episode rather than doing a one-time update until the end of the episode or the task.

More concretely, we learn two kinds of policies in the ISAR algorithm, the interaction policy and the adaptive policy. We define the interaction policy using a function $f_\theta$ with parameters $\theta$. We start by initializing $f_\theta$ with  $\theta_0$ and conduct a $N$-step interaction using $f_{\theta_0}$,  which generates a trajectory segment $(s_0, a_0, r_0 ,..., s_N)$, denoted as $\mathcal{S}_0$. Similarly, the interaction policy for trajectory segment $\mathcal{S}_i$ is denoted as $f_{\theta_i}$. We calculate the interaction loss $\mathcal{L}_{int}(f_{\theta_i})$ for $\mathcal{S}_i$ and update the policy $f_{\theta_i}$ using one gradient update with learning rate $\alpha$: 
\begin{equation}
\theta_{i+1} = \theta_{i} - \alpha \nabla_{\theta_{i}}\mathcal L_{int}(f_{\theta_i})
\end{equation}
$\mathcal{L}_{int}$ is similar to the regular $\mathcal{L}_{episode}$ that incentivizes the agent to explore the policy to reach the target. Additionally, we consider an adaptive policy in our approach. This policy learns the rules for self-adaptive exploration for the remainder of the episode. For clarity, we define the adaptive policy using the same function $f_\phi$ with different parameters $\phi$. Therefore, $f_\theta$ and $f_\phi$ share the same network structure and can exchange weights. Note that, at the beginning of the episode, we initialize $\theta_0$ with $\phi_0$. We assume an adaptive-update process with a step size of $K$, which means we collect $K$ trajectory segments $\mathcal{S}_{i\sim i+K-1}$ for adaptation policy update. The adaptation loss $\mathcal{L}_{adapt}$ is defined as the sum of the interaction losses $\mathcal{L}_{int}$ for trajectory segments $\mathcal{S}_{i\sim i+K-1}$. Therefore, the objective function of the adaptive policy $f_{\phi_i}$ is defined as follows:
\begin{equation}
\mathop{min}\limits_{\theta} \sum_{i}^{i+K-1} \mathcal L_{int}(f_{\theta_i}) = \sum_{i}^{i+K-1} \mathcal L_{int}(f_{\theta_{i-1}-\alpha \nabla_\theta \mathcal{L}_{int}(f_{\theta_{i-1}})})
\end{equation}
We update the adaptive policy $f_{\phi_i}$ using one gradient update with learning rate $\beta$:
\begin{equation}
\label{eqn:eq3}
\phi_{i+1} = \phi_i - \beta \nabla_\phi \sum_{i}^{i+K-1} \mathcal L_{int}(f_{\theta_i})
\end{equation}
According to Eq. \ref{eqn:eq3}, we note that the adaptive policy is trained by optimizing the performance of the interaction policy $f_{\theta_i}$ with respect to $\theta_i$ within the set trajectory segments $\mathcal{S}_{i \sim i+K}$. The intuition behind this approach is that by interacting a short trajectory with a small number of gradient steps, we estimate the agent's overall performance periodically within the episode and refine the navigation policy using adaptive loss and the interaction loss. As such, the agent can learn not only the sequence of actions for a specific trajectory but also become broadly applicable for the future exploration steps. This self-adaptive process enables the agent to dynamically adjust its exploration strategy which accelerates the learning process. We then update the parameters of $f_{\theta_{i+K}}$ with $\phi_{i+1}$ and repeat the incremental update process until the end of the episode. The complete algorithm for ISAR is outlined in Algorithm \ref{alg:alg1}.

\begin{algorithm}
\caption{ISAR}
\begin{algorithmic}[1]
\Require $\alpha, \beta$: learning rate
\Require $N$: trajectory segment length
\Require $K$: meta-update step size
\State randomly initialize $\theta, \phi$
\State $j \gets 0$
\While{not converged}
\State $i \gets 0$
\State $t \gets 0$
\State $\theta_i \gets \phi_j$
\While{not done}
\State $\theta_i \gets \phi_j$
\State Take action $a$ sampled from $f_{\theta_i}(s_t)$ 
\State $j \gets j+1$
\If{$t$ is divisible by N} 
\State $\theta_{i+1} = \theta_{i} - \alpha \nabla_{\theta_{i}}\mathcal L_{int}(f_{\theta_i})$
\State $i \gets i+1$
\EndIf
\If{$t$ is divisible by $K \times N$} 
\State $\phi_{j+1} = \phi_j - \beta \nabla_{\phi_j} \sum_{i}^{i+K} \mathcal L_{int}(f_{\theta_i})$
\State $\theta_i \gets \phi_{j+1}$ 
\State $j \gets j+1$
\EndIf
\EndWhile
\EndWhile
\State \textbf{return} $\theta, \phi$
\end{algorithmic}
\label{alg:alg1}
\end{algorithm}

\begin{figure*}[t!]
\centering
\includegraphics[width=\linewidth]{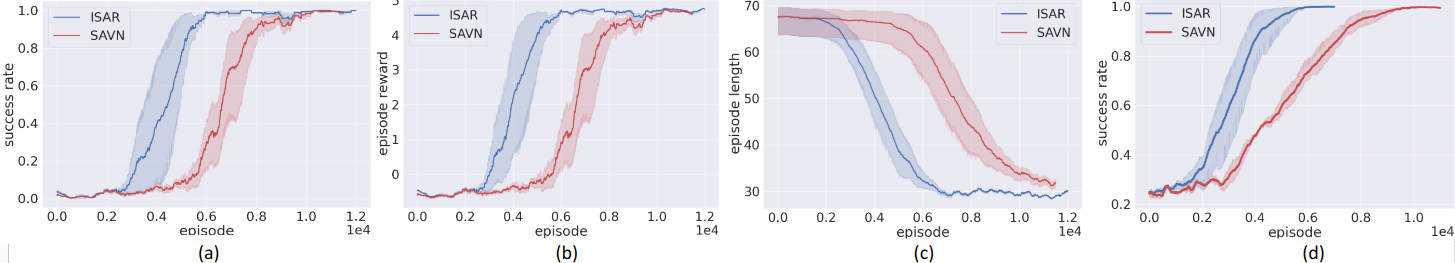}
\caption{Figure (a), (b), (c) illustrate the learning process of the ISAR and SAVN models in the subarea of AirSim urban environment. Figure (d) shows the training process of the meta-policy.}
\label{fig7}
\end{figure*}

\section{Experiments}

In this section, we evaluate the performance of the proposed methodologies for the urban navigation task in the AirSim environment. We seek answers to the following questions: (1) Can ISAR accelerate the learning process for the navigation task in comparison to traditional meta reinforcement learning? (2) Does the meta curriculum training scheme improve the learning efficiency for large-scale urban navigation? (3) Does the meta-training policy exhibit adaptation ability to new environments? For this experiment, we train the network with PyTorch using multi-process \cite{Paszke2019PyTorchAI} on NVIDIA GeForce RTX 3090.

\subsection{Learning to Navigate using ISAR}

This experiment aims to validate the efficacy of the ISAR algorithm in terms of convergence speed compared to MRL. The baseline used for comparison in this study is the \textbf{Self Aaptive Visual Navigation (SAVN)} \cite{Wortsman2018LearningTL} model. SAVN is an implementation of traditional MRL for the indoor navigation task. It optimizes the policy through the meta-loss without incremental policy updates. SAVN has been used as the base model in several recent works \cite{Luo2020AFS}\ \cite{Li2019UnsupervisedRL} \cite{Mayo2021VisualNW}.

In the experiment setup, we choose a sub-region in the AirSim urban environment as the task environment. To ensure a fair comparison, both models are trained to learn to navigate from scratch to the same destination. We use the same A3C \cite{Volodymyr} architecture as the foundational model for both ISAR and SAVN algorithms. The training process continues until convergence, indicating that the agent is capable of reaching the destination from any randomly selected starting position in the environment. We set the trajectory segment length ($N$ in Algorithm \ref{alg:alg1}) to $5$ and the adaptive-update step size ($K$ in Algorithm \ref{alg:alg1}) to $3$. The maximum episode length is set to $70$, meaning that if the target is not reached within $70$ steps, the episode will be forced to terminate. We use SGD for interaction policy updates, and Adam \cite{Kingma2014AdamAM} for adaptation policy updates.

We replicate the experiment with random seeds and evaluate the learning performance of both models in terms of episode length, average success rate and episode reward. Figure. \ref{fig7} (a), (b), (c) compare the learning process of different models. The results show that our ISAR algorithm demonstrates a significant improvement in convergence speed compared to SAVN. The ISAR model reaches convergence in approximately $5K$ learning episodes, which is $50\%$ less than the $10K$ episodes required by SAVN. Both models achieve similar average episode lengths at convergence.

\subsection{Learning to Navigate with Meta-Curriculum Training}

In this section, we present the meta-curriculum training scheme for the low-altitude urban navigation task. 

\noindent \textbf{Meta-Training.} In this stage, our objective is to train a meta-agent to `learn to learn to navigate', which can then be quickly transferred to the target environment. For meta-training, we set the agent's altitude at $300$ meters and randomly select subareas within this level as meta-tasks. Each subarea spans an area of $150m \times 150m$. In this study, we train the meta-agent with $15$ meta-tasks and implement both ISAR and SAVN for meta-training. As illustrated in Figure. \ref{fig7} (d), Our ISAR model converges in approximately $40\%$ fewer episodes ($6K$) compared to the SAVN meta-reinforcement learning model ($10K$). This result further validates the efficacy of the incremental update procedure and adaptive policy in the ISAR algorithm.

\begin{figure}[t!]
\centering
\includegraphics[width=\linewidth]{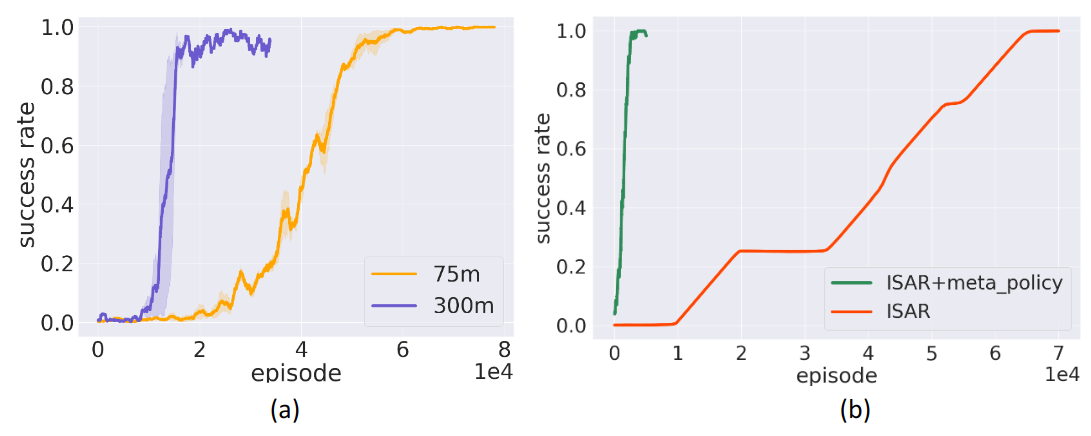}
\caption{(a) Learning process at altitudes of $300m$ and $75m$. (b) The comparison between the transfer process of meta-agent and the standard agent that learns from scratch.}

\label{fig8}
\end{figure}

\noindent \textbf{Hierarchical Curriculum Training.} In this phase, we aim to adapt the meta-agent to an lower altitude of $15$ meters. Before presenting the details, we conduct an exploration of the learning expense at different levels by training the agent to navigate from scratch at altitudes of $300m$ and $75m$. As shown in Figure. \ref{fig8} (a), the difficulty of the navigation task increases significantly as the altitude decreases. It takes around $5$ times more learning iterations to learn the policy at the $75m$ level than at the $300m$ level due to the increasing complexity of the environment. Based on this observation, we implement curriculum training by gradually introducing increasingly complex tasks to the agent in a hierarchical structure, starting from high altitude and progressing to lower altitude.

\begin{figure}[t!]
\centering
\includegraphics[width=3in]{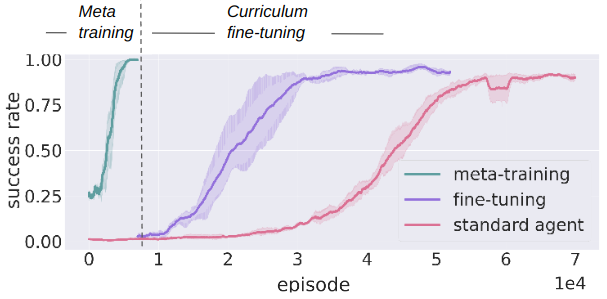}
\caption{The illustration of the meta-training phase, curriculum fine-tuning phase, and the training process of the standard agent that learns directly at target level without meta-policy.}

\label{fig10}
\end{figure}

Specifically, we initialize the agent with the meta-policy and conduct training at the altitude of $75$ meters to navigate the entire environment. The length of an exploration episode is set to $120$ steps. We then lower the drone to $45$ meters and continue training. During this process, we allows the agent to query the higher level policy at same location. Finally, we set the agent at $15$ meters level and learn to navigate to the same destination. Figure. \ref{fig10} shows our two-stage training process. For comparison, we train another standard agent which is directly placed at the target level and learns from scratch  without the meta-policy. The results demonstrate the effectiveness of the meta-curriculum training scheme in accelerating the training speed. As illustrated in Figure. \ref{fig10},  the total number of episodes required for convergence of meta-training and curriculum fine-tuning is approximately $35K$, whereas the standard agent require at least $60K$ episodes on average to achieve the same performance. An example of a navigation path is illustrated in Figure \ref{fig11} (a).


\subsection{Learning to Navigate in Unseen Environment}

\begin{figure}[t!]
\centering
\includegraphics[width=3.4in]{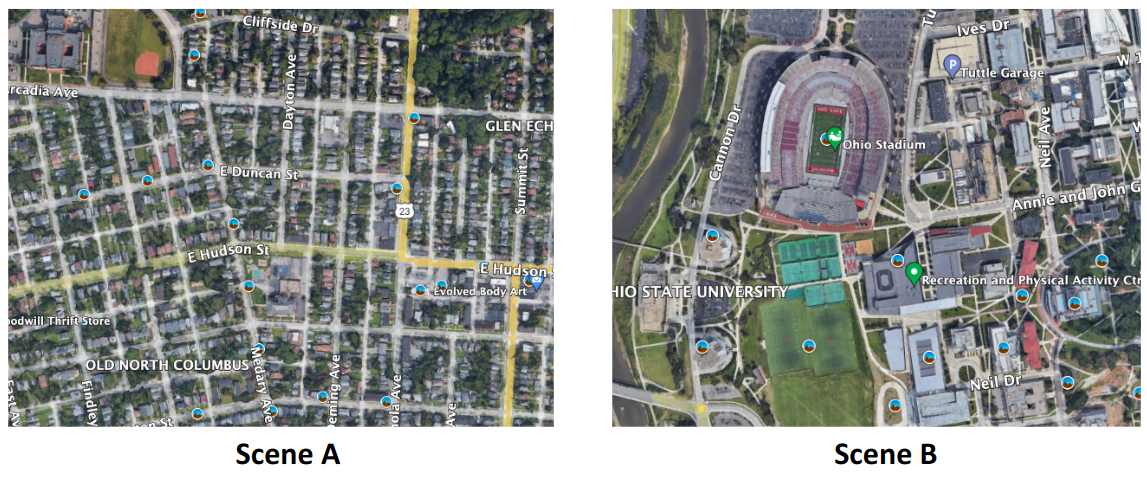}
\caption{The demonstration of real world environment. We conduct meta-training in scene A and transfer the meta-policy to scene B.}
\label{fig9}
\end{figure}

Our objectives for this experiment are to: (1) deploy the proposed approach in the real world settings; and (2) test the transfer ability of the meta-training policy to novel environments. We build an environment using Google Maps aerial photography that consists two different scenes (see Fig. \ref{fig9}). Scene A covers a residential area, whereas scene B spans the university campus with distinctive landscape. We don't include obstacles in this real world environment due to limited access to the real 3D data. 

\begin{figure}[t!]
\centering
\includegraphics[width=3.4in]{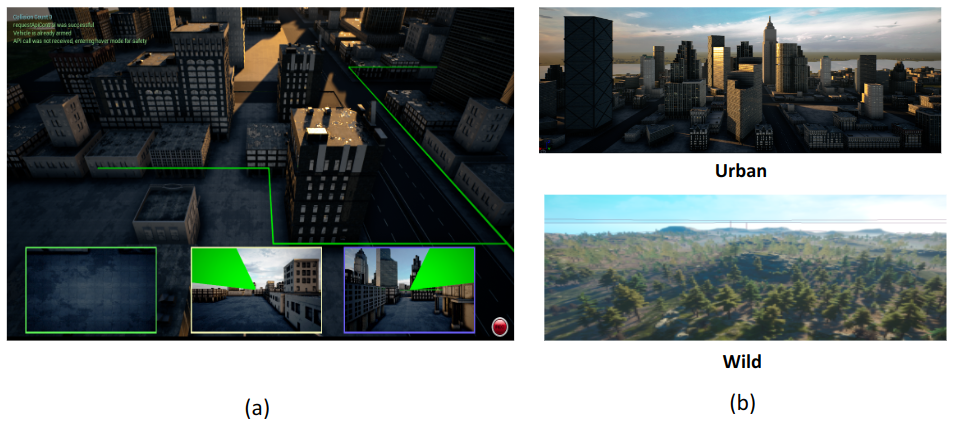}
\caption{(a) The demonstration of a navigation path. (b) The urban and wild scenes in the simulation environment.}

\label{fig11}
\end{figure}

We first conduct meta-training in scene A with $25$ meta-tasks. Then, we transfer the meta-policy for navigation in scene B through fine-tuning. For comparison, we train another agent from scratch without the meta-policy in scene B with the same destination. Figure. \ref{fig8} (b) shows the transfer process of the meta-agent and the normal agent. The result demonstrates the efficacy of our methods in real scenarios. Moreover, the meta-training model exhibits significantly rapid transfer ability to unseen environments, converging within around $2K$ episodes and achieving a convergence speed $25$ times faster than learning from scratch ($50K$ episodes). However, it has limitations in certain situations. We observe that if the new environment is significantly different from the meta-training environment, such as from urban to wild (see Fig. \ref{fig11} (b)), the meta-training exhibits poor transfer performance.

\subsection{Ablation Discussion}

In this section, we perform an ablation discussion to gain further insights into our proposed approaches. 

\noindent \textbf{Impact of trajectory segment length N.} The trajectory segment length $N$ refers to the number of interactions the agent takes before updating its interaction policy each time. We observe that too short trajectories provide insufficient information to update the interaction policy, while excessively long trajectories reduce efficiency. Notably, the optimal value of $N$ is similar to the average path length.

\noindent \textbf{Impact of meta-update step size K.} Our ISAR algorithm updates the adaptation policy with $K$ trajectory segments each time. We observe that updating the adaptive policy too frequently with insufficient information or too infrequently can reduce efficiency. The optimal value of $K$ strikes the best balance between exploration and learning update.

\noindent \textbf{Impact of size and number of the meta-task.} When larger and more meta-tasks are used during meta-training, the meta-agent gains a broader view and more comprehensive understanding of the environment, which results in better generalization ability. However, training with more meta-tasks would take longer. Therefore, the balance between training and transfer efficiency should be considered.

\section{Conclusion}

In this paper, we present a meta-curriculum training strategy to facilitate the learning process of large-scale low-altitude urban navigation task. By presenting progressively more complex environments to the agent, curriculum training allows the agent to quickly refine its policy to the target task. Additionally, we propose an incremental self-adaptive reinforcement learning (ISAR) algorithm to accelerate the learning speed of traditional meta-reinforcement learning. ISAR leverages the incremental update procedure, interaction loss, and adaptation loss to facilitate self-adaptive exploration within the episode. Our experiments show that the ISAR model outperforms traditional MRL approaches in terms of convergence speed for long-range navigation tasks. Moreover, we not only demonstrate the efficacy of our training scheme for large-scale environment navigation tasks, but also show that the meta-training policy has the ability to rapidly adapt to novel environments.

\section{Acknowledgement}


This work was supported in part by the U.S. Army Research Office under Grant AWD-110906.

{
	\begin{spacing}{1.17}
		\normalsize
		\bibliography{ISPRSguidelines_authors} 
	\end{spacing}
}

\end{document}